\documentclass[lettersize,journal]{IEEEtran}
\usepackage{amsmath,amsfonts}
\usepackage{algorithmic}
\usepackage{algorithm}
\usepackage{array}
\usepackage[caption=false,font=normalsize,labelfont=sf,textfont=sf]{subfig}
\usepackage{textcomp}
\usepackage{stfloats}
\usepackage{url}
\usepackage{verbatim}
\usepackage{graphicx}
\usepackage{cite}
\usepackage{amsmath}
\usepackage{threeparttable}
\usepackage{booktabs}
\usepackage{multirow}
\usepackage{cite}
\begin{document}

\title{Bridging the View Disparity Between Radar and Camera Features for Multi-modal Fusion 3D Object Detection}

\author{Taohua~Zhou,
	Yining~Shi,
	Junjie~Chen~\IEEEmembership{Member,~IEEE},
	Kun~Jiang,
	Mengmeng~Yang,
	and~Diange~Yang
\thanks{T. Zhou, Y.Shi, J.Chen, K.Jiang, M. Yang, and D. Yang are with the State Key Laboratory of Automotive Safety and Energy, School of Vehicle and Mobility, Tsinghua University, Beijing, China (e-mail: ydg@mail.tsinghua.edu.cn; jiangkun@tsinghua.edu.cn; junjiec@tsinghua.edu.cn).}
\thanks{Manuscript submitted August 22, 2022}}

\markboth{Journal of \LaTeX\ Class Files,~Vol.~14, No.~8, August~2021}%
{Shell \MakeLowercase{\textit{et al.}}: A Sample Article Using IEEEtran.cls for IEEE Journals}


\maketitle

\begin{abstract}
Environmental perception with the multi-modal fusion of radar and camera is crucial in autonomous driving to increase accuracy, completeness, and robustness. This paper focuses on utilizing millimeter-wave (MMW) radar and camera sensor fusion for 3D object detection. A novel method that realizes the feature-level fusion under the bird's-eye view~(BEV) for a better feature representation is proposed. Firstly, radar points are augmented with temporal accumulation and sent to a spatial-temporal encoder for radar feature extraction. Meanwhile, multi-scale image 2D features which adapt to various spatial scales are obtained by image backbone and neck model. Then, image features are transformed to BEV with the designed view transformer. In addition, this work fuses the multi-modal features with a two-stage fusion model called point-fusion and ROI-fusion, respectively. Finally, a detection head regresses objects category and 3D locations. Experimental results demonstrate that the proposed method realizes the state-of-the-art~(SOTA) performance under the most crucial detection metrics——mean average precision~(mAP) and nuScenes detection score~(NDS) on the challenging nuScenes dataset.
\end{abstract}


\section{Introduction}

\IEEEPARstart{I}{ntelligent} vehicles are usually equipped with multiple sensors to enhance the environment perception ability for safety. While LiDAR perception methods have obtained excellent performance with precise 3D measurement, the expensive cost of LiDAR makes it hard to be applied for large-scale deployment on intelligent vehicles. Compared with LiDAR, camera and radar units are easier to be accepted, and radar-camera fusion has been widely applied in the Advanced Driving Assistance System~(ADAS) for intelligent vehicles. MMW radar can directly measure objects' relevant position and velocity, be robust to adverse weather conditions, and be at a low cost~\cite{cai2021machine}. However, radar point clouds lack semantic information and can not avoid clutter from the environment. Camera images own much semantic information but are quite sensitive to light and occlusions. Therefore these two sensors can complement each other and exploit the advantages of both sensors. However, measurements from different sources are represented in heterogeneous space, multi-modal fusion detection is more complex than single-sensor perception. \par

In general, there are three kinds of radar-camera fusion schemes to solve fusion-based object detection. The first fusion scheme is output fusion which fuses the single-sensor detection output for a higher-confidence result with the D-S evidence theory~\cite{chavez2015multiple} or fusion filter algorithm~\cite{chavez2014fusion}. This method can not exploit raw sensor data adequately and is quite dependent on the probability modeling of single sensor results. The second fusion framework is raw data fusion which usually projects radar points to the image plane. Then projected radar points provide hints of the region of interest~(ROI) for further processing~\cite{nabati2019rrpn, alessandretti2007vehicle}. 
This kind of fusion algorithm is more suitable for boosting 2D image tasks rather than 3D detection.
The third one, called feature fusion, exploits Deep Neural Networks~(DNN) to generate and fuse high-dimensional features from radar data and images. The mainstream paradigm is projecting radar points onto the image plane with elaborate rules for feature extraction. Feature fusion can fully exploit the intermediate features from raw sensor data and be designed to solve various regression problems, including 2D image bounding box estimation, 2.5D (2D image bounding box and depth estimation), and 3D object detection~\cite{nabati2020radar, chang2020spatial, nabati2021centerfusion}. \par 
However, as camera captures semantic information in the front view and radar obtains spatial information in the 3D space, direct projecting through a calibration matrix ignores the 
view discrepancy between different sensors. This kind of fusion method causes the loss of high-dimension sensory information for 3D perception, which consequently impacts the fusion effect. Lim et al. were also aware of this problem and devised an Inverse Projection Mapping~(IPM) method to realize image information presentation under the same space as radar~\cite{lim2019radar}. 
However, since direct IPM causes distortion and loss of image features, realizing spatial-temporal synchronization at the raw data stage may not be the most suitable approach for radar-camera fusion-based detection. \par
According to the discussion above, it is necessary to design a suitable representation of fusion features considering the characteristics of radar points and image information to avoid information loss. The success of visual perception works under the bird's-eye view~\cite{philion2020lift, huang2021bevdet}, also called the top-down view, gives us some heuristic information. We consider representing the cross-view features from multi-modal data for spatial-temporal synchronization. Besides, When we visualize radar point clouds and object 3D bounding box under the bird's-eye view, we find that radar point clouds are distributed among the box edges. We intend to use this geometry constraint to enhance the fusion performance. Therefore a two-stage fusion method enhances information interaction when extracting features and regressing the region of interest. Considering the sparse and clutter of radar points, a temporal-spatial feature extractor is introduced to process radar data.\par

Hence this paper will introduce a cross-view feature-level fusion framework called RCBEV, to realize 3D object detection with radar and camera data. With the proposed radar branch module, efficient features are extracted by overcoming the sparsity and clutter of radar points. Through a designed multi-scale view transformer, cross-view image features are obtained. Then a two-stage fusion scheme, including point-fusion and ROI-fusion, is realized. Finally, 3D detection results are acquired with an anchor-free regression head.
The main contributions of our work lie in three parts:\par
1) A novel 3D object detection network for feature-level radar-camera fusion is proposed. To the best of our knowledge, we are the first to bridge the view disparity between multi-modal features for radar-camera fusion detection.  Cross-view fusion features are exploited to avoid information loss, and a two-stage fusion module is designed to exploit multi-modal features adequately.  \par
2) A temporal-spatial encoder is designed to process radar data and deal with the sparsity and clutter of radar point clouds. Besides, considering the geometry constraints provided by radar, high-dimensional heatmaps  are generated to enhance the spatial information regression compared with vision-only networks.\par
3) The effectiveness of RCBEV is verified on the nuScenes dataset. Extensive experiments are carried out to prove the overall effectiveness and model robustness under special situations. RCBEV realizes the SOTA  under the camera-radar modality of the nuScenes detection leaderboard.\par

The rest of this paper is organized as follows. Section~\ref{sec-overview} presents an overview of previous studies. Section~\ref{sec-system} introduces the overall architecture of RCBEV. Then Section~\ref{sec-method} explains the detailed algorithm design. Experiments are conducted and analyzed using the  public autonomous driving dataset nuScenes in Section~\ref{sec-experiment}. Finally, the conclusion of this proposed work is summarized in Section~\ref{sec-conclusion}. \par

\section{Related Work}\label{sec-overview}
\subsection{Single-sensor 3D Object Detection}
With the precise 3D spacial information measured by LiDAR, related 3D object detection methods are firstly developed. Two-stage LiDAR 3D detectors such as PointRCNN~\cite{shi2019pointrcnn}, and PV-RCNN~\cite{shi2020pv}, single-stage detectors such as Voxelnet~\cite{zhou2018voxelnet}, and PointPillars~\cite{lang2019pointpillars}, and anchor-free detectors such as CenterPoint~\cite{yin2021center}, and Object DGCNN~\cite{wang2021object} have made a breakthrough. \par 
Due to the high cost of LiDAR, camera-only 3D perception also followed close on another. Unlike LiDAR sensors, lack of depth causes the difficulty to image 3D  detection. The solutions are mainly divided into the following types. The earliest works try to import geometry prior information and camera model to get 3D information~\cite{mousavian20173d}. However, these proposals are easily impacted by occlusion, truncation, and height estimation errors. The second attempt is to utilize image features with additional depth information to produce pseudo LiDAR point clouds and  get the final detections using point cloud processing networks~\cite{weng2019monocular}. The third type extends the 2D image detector with additional 3D regression branches, such as FCOS3D~\cite{wang2021fcos3d}, and CenterNet~\cite{zhou2019objects}.  The fourth design is the transformer-based detection method with learnable object queries to get 3D object information such as DETR3D~\cite{wang2022detr3d} and PETR~\cite{liu2022petr}. The last and the most concerned ones project image-view features to BEV through a view transformer, such as Imvoxelnet~\cite{rukhovich2022imvoxelnet}, BEVDet~\cite{huang2021bevdet}, and M2BEV~\cite{xie2022m}. Up to now, multi-image 3D detection under BEV is the most concerned issue in this research field. \par
\subsection{Fusion-based Object Detection}
Multi-modal fusion detection is mainly developed with LiDAR and camera. Methods are divided into early fusion, such as PI-RCNN~\cite{xie2020pi}; feature fusion, such as 3d-cvf~\cite{yoo20203d}; and late fusion, such as CLOCs~\cite{pang2020clocs}. Semantic knowledge from image 2D  information is utilized to augment LiDAR point clouds to get more precise results. How to fusion multi-modal data efficiently, and how to integrate multi-view features, are still the biggest challenges in this field. \par 
As for radar-camera fusion detection, the  development progress is clearer. With the advance of image 2D detection paradigms, such as Faster RCNN~\cite{ren2015faster} and YOLO~\cite{redmon2016you}, researchers explore utilizing radar data augmentation to improve the detection performance when there exists occlusion or when in a raining environment or dark night. SAF-FCOS~\cite{chang2020spatial}, CRF-Net~\cite{nobis2019deep}, RVNet~\cite{john2019rvnet}, and other studies have acquired convincing evidences to prove that radar data can help to further promote the detection performance under these special scenarios. \par 

Moreover, with the development of 3D image object detection, CenterFusion~\cite{nabati2021centerfusion} realizes 3D object detection based on radar-camera feature fusion and proved that the result surpasses the baseline model CenterNet~\cite{zhou2019objects}. However, dealing with the sparsity and noise of radar data and utilizing the different-modal data to realize efficient feature representation when implementing fusion-based perception tasks have always been the key issues~\cite{schumann2019scene, cai2021autoplace}. \par

\begin{figure*}[!t]
	\centering
	\includegraphics[scale=0.6]{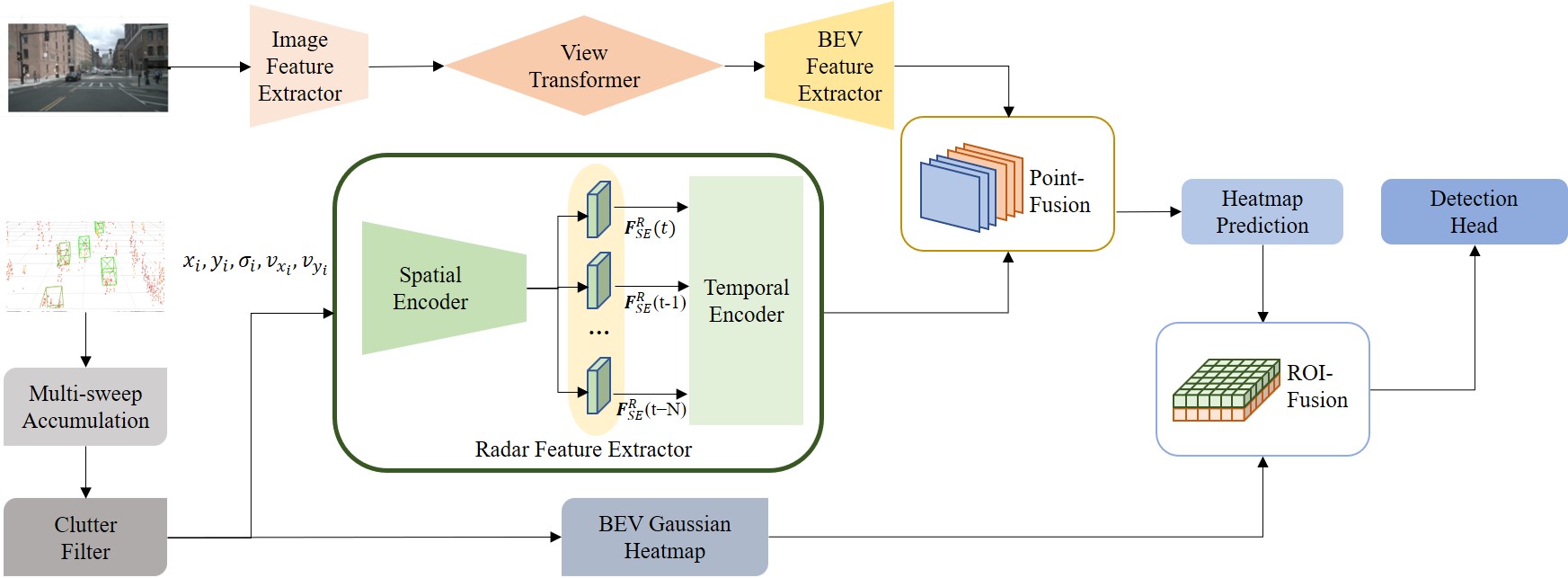}
	\caption{Overview of the proposed RCBEV for 3D object detection}
	\label{fig-frame}
\end{figure*}

\section{System Architecture}\label{sec-system}
This work aims to build a multi-modal feature fusion approach by bridging the view disparity between radar and image features to enhance the 3D object detection performance for autonomous driving vehicles. The overview of the proposed system architecture is illustrated in Fig.~\ref{fig-frame}, and the concrete model design is shown in Table~\ref{tab-NA}. \par

At first, multi-scale features under image view are extracted through a well-performed backbone network, such as Swin-Transformer~\cite{liu2021swin}, and a neck model Feature Pyramid Network (FPN)~\cite{lin2017feature}. Next, image-view features and depth estimation results are lifted to a visual frustum with a designed view transformer.  Then, features are further extracted under the BEV space. \par

Meanwhile, this paper proposes a speciﬁc scheme to  deal  with the sparsity and clutter of radar points. Firstly, preprocessing procedures are implemented on raw radar data. Multi sweeps of radar points are accumulated to overcome data sparsity, and a confidence filter is designed to cover clutter and false alarms. Secondly, a radar feature extractor is proposed. Unlike 3D outdoor measurements, radar points only provide 2D location information $x_r$ and $y_r$ with no height information. Ego-motion compensated Doppler velocity $v_x$ ,  $v_y$, and the radar cross section~(RCS) $\sigma$ which reflects signal intensity is also utilized to enrich radar features. Then the 5-dimensional data is sent to the spatial feature encoder to extract point cloud features. Then a temporal encoder is designed to overcome feature misalignment of multiple sweeps with time series prediction. \par

As for the fusion module, we design a two-stage fusion strategy for the sufficient multi-modal feature interaction. The synchronized image features and radar features  are fused at the point-fusion stage. Next, we use radar input to generate heatmaps that obey Gaussian distribution under the unified BEV space to reflect the geometry constraints of objects. Then ROI-fusion is realized with the predicted heatmaps from the upstream module and radar heatmaps. Finally, the augmented heatmaps are used to further regress categories and 3D bounding boxes with an anchor-free 3D detection head.\par
\begin{table*}[htpb!]
	\centering
	\small
	\caption{Network Architecture}
	\label{tab-NA}
	\scalebox{1.0}{
		\begin{tabular}{c|cccc}
			\hline
			\hline
			\multirow{2}{*}{Module} & 
			\multirow{2}{*}{Component} & 
			\multirow{2}{*}{Name} & 
			\multirow{2}{*}{Input Size} & 
			\multirow{2}{*}{Output size} \\
			&&&&\\
			\hline
			
			\multirow{8}{*}{Image Feature Extractor} & 
			\multirow{2}{*}{Image-view Backbone} &\multirow{2}{*}{Swin Transformer} &\multirow{2}{*}{$ H\times W\times3$} &$\frac{H}{16}\times \frac{W}{16} \times 384$\\
			&&&&$ \frac{H}{32}\times \frac{W}{32} \times 768$  \\
			& \multirow{2}{*}{Image-view Neck} &  \multirow{2}{*}{FPN}  &\multirow{2}{*}{Output of up layers} &\multirow{2}{*}{$ \frac{H}{16}\times \frac{W}{16} \times $1152} \\
			&&&&\\
			& \multirow{2}{*}{View Transformer} &\multirow{2}{*}{-} & \multirow{2}{*}{Output of up layers}& \multirow{2}{*}{$\frac{H}{16}\times\frac{W}{16}\times D\times64$} \\ 
			&&&&\\
			& \multirow{2}{*}{BEV  Feature Extractor }&\multirow{2}{*}{-} & \multirow{2}{*}{Output of up layers} & \multirow{2}{*}{$\frac{H}{16}\times \frac{W}{16}\times512$}\\
			&&&&\\
			\hline
			
			\multirow{4}{*}{Radar Feature Extractor} & 
			\multirow{2}{*}{Spatial Encoder} & \multirow{2}{*}{Modified VoxelNet}& \multirow{2}{*}{$M\times5$} &\multirow{2}{*}{$N\times\frac{H}{4}\times \frac{W}{4}\times512$} \\
			&&&&\\
			&\multirow{2}{*}{Temporal Encoder}  & \multirow{2}{*}{ConvLSTM}& \multirow{2}{*}{Output of up layers} &\multirow{2}{*}{$\frac{H}{4}\times \frac{W}{4}\times512$} \\
			&&&&\\
			\hline
			
			\multirow{6}{*}{Point-fusion~(PF)} & 
			\multirow{2}{*}{FOV Alignment}  &Downsampling &$ \frac{H}{16}\times \frac{W}{16}\times512$ &$ \frac{H}{8}\times \frac{W}{8}\times256$ \\
			 &&Upsampling&$\frac{H}{4}\times \frac{W}{4}\times512$&$\frac{H}{8}\times \frac{W}{8}\times256$ \\
			&\multirow{2}{*}{PF Operation1} & \multirow{2}{*}{Concatenation} & \multirow{2}{*}{Output of up layers}&\multirow{2}{*}{$ \frac{H}{8}\times \frac{W}{8} \times $512} \\
			&&&&\\
			&\multirow{2}{*}{PF Operation2}& \multirow{2}{*}{Convolution} & \multirow{2}{*}{Output of up layers} & \multirow{2}{*}{$\frac{H}{8}\times \frac{W}{8} \times256$} \\ 
			&&&&\\
			
			\hline
			\multirow{8}{*}{ROI-fusion~(RF)} &
			\multirow{2}{*}{PF HeatMap Predictor} & \multirow{2}{*}{Shared convolution layer} & \multirow{2}{*}{$\frac{H}{8}\times \frac{W}{8} \times256$ }& \multirow{2}{*}{$\frac{H}{8}\times \frac{W}{8} \times256$}\\
			&&&&\\
			&\multirow{2}{*}{Radar Heatmap Generator}&\multirow{2}{*}{-} &\multirow{2}{*}{$M\times6$} &\multirow{2}{*}{$\frac{H}{8}\times \frac{W}{8} \times6$}  \\
			&&&&\\
			&\multirow{2}{*}{RF Operation1} & \multirow{2}{*}{Multiply}&$\frac{H}{8}\times \frac{W}{8} \times256$ &\multirow{2}{*}{$\frac{H}{8}\times \frac{W}{8}\times256\times6$}\\
			&&&$\frac{H}{8}\times \frac{W}{8} \times6$&\\
			&\multirow{2}{*}{RF Operation2}& \multirow{2}{*}{Convolution}& \multirow{2}{*}{Output of up layers} & \multirow{2}{*}{$\frac{H}{8}\times \frac{W}{8}\times256$} \\
			&&&&\\
			
			\hline
			\multirow{2}{*}{Detection Head} & 
			\multirow{2}{*}{Regression Layers} &\multirow{2}{*}{ -} &\multirow{2}{*}{$\frac{H}{8}\times \frac{W}{8}\times256$}&\multirow{2}{*}{ $O\times10$} \\
			&&&&\\
			\hline\hline
		\end{tabular}
}
\begin{tablenotes} 
	\footnotesize 
	\item $H$ and $W$ denote the resolution of input images. $M$ is the number of input radar points. $D$ is the resolution of discrete depth estimation. $N$  denotes the time window length of the radar accumulation. $O$ denotes the number of detection results.
\end{tablenotes}  
\end{table*}
\section{Methods}\label{sec-method}
This section studies the cross-view 3D detection model fusing radar and camera data at the feature level. Our RCBEV model consists of 4 parts: radar branch, image branch, fusion branch, and detection head. The detailed design of each part is emphasized in the following content.\par
\subsection{Radar processing}\label{ssec-radar}
\begin{table}[htbp]
	\centering
	\caption{Utilized Radar information in the nuScenes dataset}
	\begin{threeparttable}
		\vspace{0.2mm}
		\scalebox{0.95}{
		\begin{tabular}{ccc}
			\toprule
			\toprule
			Information &Meaning & Unit \\ 
			\midrule
			$x_r$ &Position in front &m   \\
			$y_r$  &Position in left &m \\
			$v_x$ &Velocity  in front compensated by the ego motion&m/s \\
			$v_y$ &Velocity  in left compensated by the ego motion&m/s \\
			$\sigma$ &Radar cross section reflecting echo intensity &dB \\
			$\gamma_{vs}$ &State of cluster validity &-\\
			$\gamma_{ds}$ & State of Doppler (radial velocity) ambiguity solution&-\\
			$\gamma_{fp}$& False alarm probability of cluster &-\\
			$x_{rms}$ &Uncertainty of $x_r$ (root mean square value)&m   \\
			$y_{rms}$  &Uncertainty of $y_r$ (root mean square value) &m \\
			$v_{x_{rms}}$ &Uncertainty of $v_x$(root mean square value)  &m/s \\
			$v_{y_{rms}}$ &Uncertainty of $v_y$ (root mean square value)&m/s \\
			\bottomrule
			\bottomrule			
		\end{tabular}
	}
	\end{threeparttable}
	\label{tab-radar-info}
\end{table}
To fully exploit radar points for 3D object detection, the sparsity and clutter need to be handled, as well as suitable data structures for fusion-based detection tasks need to be constructed.  Consequently, three modules of the radar branch are specially designed: 1) data preprocessing module, 2) radar feature extractor with a spatial-temporal encoder, and 3) object heatmap generator. And the illustration of 1) and 2) is shown in Fig. 2. The utilized radar measurements $ \mathbf{Z}^{Radar}  $ are presented in~(\ref{eq-Z-radar}). The physical meanings of these symbols are shown in Table~\ref{tab-radar-info}.\par
\begin{equation}
\label{eq-Z-radar}
	\begin{aligned}
	\mathbf{Z}^{Radar} = &\left\lbrace x_r,y_r,v_x,v_y,\sigma,\gamma_{vs},\gamma_{ds},\gamma_{fp}, a_{rms}\right\rbrace \\
	a=&\left\lbrace x_r,y_r,v_x,v_y\right\rbrace 
	\end{aligned}
\end{equation}
\par

In the data preprocessing module, we realize a multi-sweep integrator to increase the density of points. Based on sensor calibration parameters $ \mathbf{T}^{ego(t-k)}_{radar} $ and $ \mathbf{T}^{ref(t)}_{ego} $, global localization information $ \mathbf{T}^{ego(t)}_{global} $ and $ \mathbf{T}^{global}_{ego(t-k)} $, and timestamp of sensor data recording time, radar points $ \mathbf{Z}^{radar}(t-k) $ at previous timestamp $ t-k $ can be unified at current frame $ t $ under the reference coordinate system as $\mathbf{Z}^{ref}(t)$. Here we set the front radar coordinate system as the reference coordinate system. The mathematical formulation of the above course is listed as

\begin{equation}
	\label{eq-radar1}
	\begin{aligned}
		\mathbf{Z}^{ref}(t) &=Trans(\mathbf{Z}^{radar}(t-k))\\ &=\mathbf{T}^{ref(t)}_{ego}\mathbf{T}^{ego(t)}_{global}\mathbf{T}^{global}_{ego(t-k)}\mathbf{T}^{ego(t-k)}_{radar}\mathbf{Z}^{radar}(t-k)
	\end{aligned} 
\end{equation}

In ~(\ref{eq-radar1}), $\mathbf{T}^{X}_Y$ denotes the transformation matrix from the Y coordinate system to the X coordinate system. It is made up of rotation matrix $\mathbf{R}^X_Y$ and translation vector $\mathbf{t}^X_Y$, as it is shown in~(\ref{eq-matrix}). \par

\begin{equation}
	\label{eq-matrix}
	\begin{aligned}
		\mathbf{T}^X_Y = [\mathbf{R}^X_Y|\mathbf{t}^X_Y] =\left[
		\begin{array}{cc}
			\mathbf{R}^X_Y& \quad\mathbf{t}^X_Y \\\\ \mathbf{0} &\quad 1
		\end{array}
		\right]
	\end{aligned} 
\end{equation}

The clutter filter in (\ref{eq-cfilter}) is designed to evaluate the confidence of radar points and ensure the training data quality. According to their dynamic properties $\gamma_{ds}$, reflected intensities $\sigma$, probabilities of being objects $\gamma_{vs}$, and probabilities of  false alarms $\gamma_{fp}$, the points which are not satisfied with the conditions are filtered out. \par 
\begin{equation}
	\label{eq-cfilter}
	CF(\mathbf{Z}^{radar}(t)) = Filter(\sigma,  \gamma_{vs}, \gamma_{ds}, \gamma_{fp})
\end{equation}
\par 
According to (\ref{eq-radar1}) and (\ref{eq-cfilter}), the whole preprocessing, including sequential accumulation and clutter filter, is expressed in~(\ref{eq-preprocess}), where $ N $  denotes the time window size for the multi-sweep accumulation. \par
\begin{equation}
	\label{eq-preprocess}
	\mathbf{I}^{radar}(t) = \sum_{k=0}^{N}{Trans\left( CF\left( \mathbf{Z}^{radar}(t-k)\right) \right) }
\end{equation}

\begin{figure}[thpb]
	\centering
	\includegraphics[width=\linewidth]{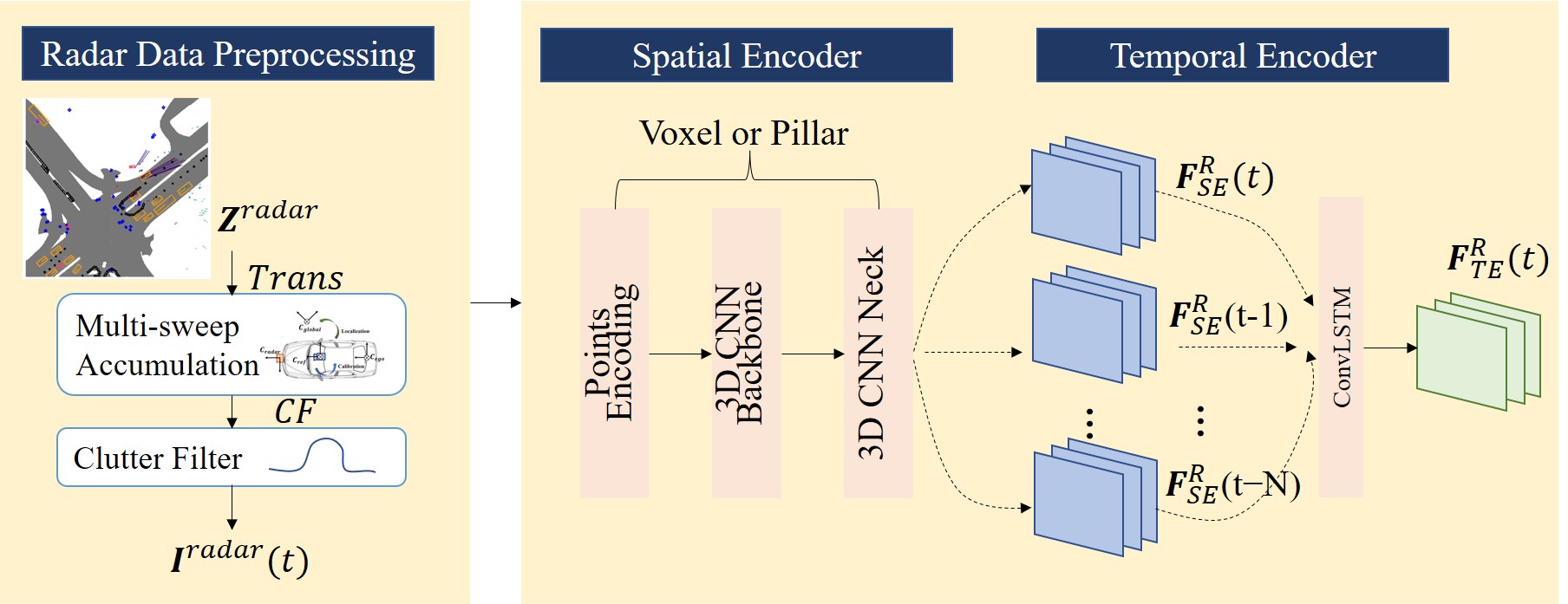}
	\caption{\label{fig-radar}Illustration of radar data processing}
\end{figure}
\par

After getting the input data $ \mathbf{I}^{radar} $, our radar feature extractor is constructed to extract features at a higher dimension. Five-dimensional information from radar point clouds, including radial distance $x_r$, lateral distance $y_r$, ego-motion compensated Doppler velocity $v_x$, $v_y$, and the RCS value $\sigma$ form the input. As the most efficient structures of point encoders include voxel encoder and pillar encoder, we use the backbone model from VoxelNet~\cite{zhou2018voxelnet} or PointPillars~\cite{lang2019pointpillars} to extract corresponding features $\mathbf{F}^{R}_{SE} $. As the voxel encoder obtains more ideal results, we finally adopt this form. \par
Although the features of different sweeps are unified under the current timestamp and the reference system, there still exist slight misalignments between different sweeps, which impact the detection performance. Hence a temporal encoder with the structure of ConvLSTM~\cite{shi2015convolutional} is built to refine the features of $ \mathbf{F}^{R}_{SE} \left( t-k\right) , k=0,1,...,N$. ConvLSTM is a kind of Long Short-Term Memory networks which extracts 2D  sequential features and realizes temporal prediction. The whole process of the radar feature extractor, including spatial encoder $SE$ and temporal encoder $TE$  is denoted as~(\ref{eq-st-encoder}). As the output, $ \mathbf{F}^{R}_{TE}(t)  $ is sent to fuse with image-branch processing results. In addition, the generation of radar heatmaps is mentioned in Sec.~\ref{subs-fusion}.\par 
\begin{equation}
	\label{eq-st-encoder}
	\begin{aligned}
		\mathbf{F}^{R}_{SE}(t-k) =& SE\left( Trans\left( CF\left( \mathbf{Z}^{radar}(t-k)\right) \right) \right)  \\
		\mathbf{F}^{R}_{TE}(t) =& TE(Concat(\mathbf{F}^{R}_{SE}(t), ...\mathbf{F}^{R}_{SE}(t-N)) \\
		k=&0,1,...,N\\
	\end{aligned}	
\end{equation}
\subsection{Image processing}\label{ssec-image}
The image branch provides important semantic information in RCBEV and is divided into three parts: 1) image-view feature extractor, 2) cross-view feature transformer, and 3) BEV feature extractor. \par

The image-view feature extractor is designed to extract multi-scale image features suitable for detection. A good backbone network with a suitable neck model is essential to visual perception tasks. Thus multi-scale features $\mathbf{F}^I_{img} $ with different field of view are obtained as (\ref{eq-img-1}).\par
\begin{equation}
	\label{eq-img-1}
	\mathbf{F}^I_{img} =  Neck(Backbone(\mathbf{I}^{image}))
\end{equation}	
\par 
Then a cross-view feature transformer is  designed to conduct image features transformation from a front-view perspective to a bird's-eye-view perspective. Considering that the direct IPM method, which projects image features to BEV with a homography matrix, is based on an assumption of flat ground. It often causes distortion of dynamic objects in the real environment. Furthermore, it is easily impacted by the turbulence of vehicles or roads. Therefore we adopt another scheme that lifts the image-view features with depth estimation and splats to the reference system. \par 
We use image-view features to build vision frustum $ \mathbf{VF} $ which reflects semantic feature  $\mathbf{F}^I_{img} $ and concrete depth distribution $\mathbf{F}^I_{Depth} $ through an exterior product operation.  Here $\mathbf{F}^I_{Depth} $ is obtained through a depth estimation operation from $\mathbf{F}^I_{img} $ with a softmax function. Then with extrinsics $ \mathbf{E}  $ and intrinsics $ \mathbf{I} $ from the camera model, features expressed in vision frustum are splatted to the reference space under BEV, where radar and image features will be synchronized. \par 
\begin{equation}
	\label{eq-img-2}
	\begin{aligned}
		\mathbf{F}^I_{Depth} = &Softmax(\mathbf{F}^I_{img})\\
		\mathbf{VF} = & \mathbf{F}^I_{img} \otimes \mathbf{F}^I_{Depth}\\
	\end{aligned}
\end{equation}	
\par
\begin{figure}[thpb]
	\centering
	\includegraphics[width=\linewidth]{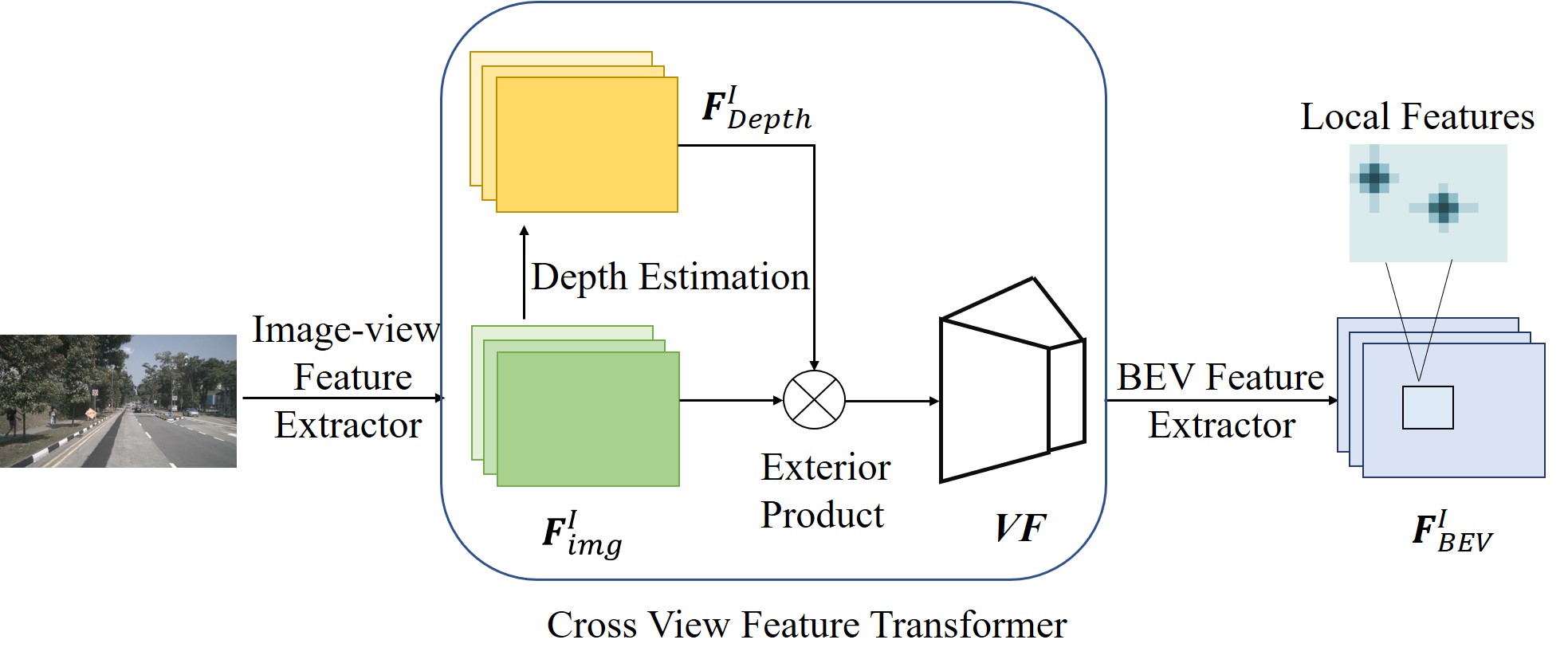}
	\caption{\label{fig-image}Illustration of image data processing}
\end{figure}

 
The BEV feature extractor is arranged to further extract features $ \mathbf{F}^I_{BEV} $ to fit the information representation under the top-down view. We use the basic layers of Resnet and FPN to operate this module. The progress of image-view feature extraction, view transformation, and BEV feature extraction is expressed in Fig.~\ref{fig-image}. \par
\begin{equation}
	\label{eq-img-eocoder}
		\mathbf{F}^I_{BEV}= BEV\left( \mathbf{E} \left( \mathbf{I}^{-1}\left( \mathbf{VF} \right) \right) \right) 
\end{equation}

\subsection{Fusion module}\label{subs-fusion}
\begin{figure}[!t]
	\centering
	\includegraphics[width=\linewidth]{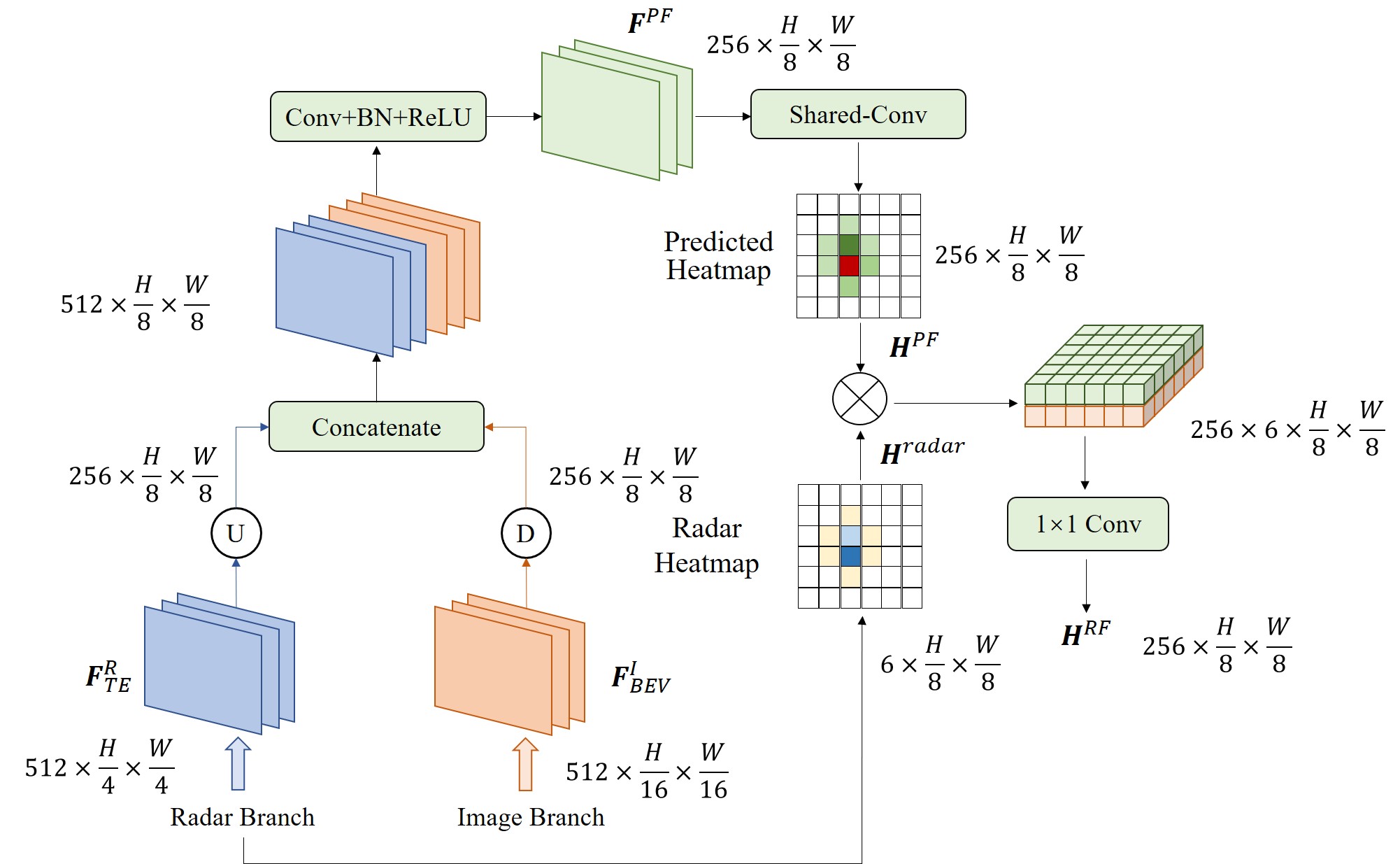}
	\caption{\label{fig-fusion}The structure of the fusion network}
\end{figure}

As shown in Fig.~\ref{fig-fusion}A two-stage R-C fusion method, a two-stage R-C fusion method is put forward to realize a sufficient feature interaction. Based on the operation order in the forward propagation, the two fusion modes are called point-fusion and ROI-fusion, respectively.\par
The first module fused the different sensor-branch features $ \mathbf{F} ^R_{TE} $ in (6) and $ \mathbf{F} ^I_{BEV} $in (9), which are both regarded as the high-dimensional encoding forms of point clouds. Hence this fusion stage is called point-fusion. As the field of views between two branches are different, and radar features are still sparser than image features, downsampling and upsampling operation are employed on image and radar features separately to unify the resolutions of different feature maps. After getting the same resolution, two-branch features are fused with a concatenation operator. Considering feature misalignments may exist caused by the inaccurate spatial transformation, a classical convolution layer is applied to fuse the image features $ \mathbf{F}^I_{BEV} $ and radar features $ \mathbf{F}^{R}_{TE} $. $ \mathbf{F}^{PF} $. The above progress is formulated in~(\ref{eq-point-fusion}).\par
\begin{equation}
	\label{eq-point-fusion}
	\mathbf{F}^{PF}=Conv\left( Concat\left( Up\left( \mathbf{F}^{R}_{TE}\right) , Down\left( \mathbf{F}^I_{BEV} \right) \right) \right) 
\end{equation}
\par 
As for the second-stage fusion, it is realized in the form of object heatmaps, which reflect objects  position and confidence under the bird's-eye view. The outputs of the first fusion stage $ \mathbf{F}^{PF} $ are sent to a shared convolution layer to predict object heatmaps $  \mathbf{H}^{PF} $ of different categories. \par 
Radar heatmaps $\mathbf{H}^{radar}(x,y)$ are generated from $\mathbf{I}^{radar}$ in Sec~\ref{ssec-radar} with a two-dimensional Gaussian distribution. 
When generating the multi-channel heatmaps, raw data  including position $x_r$ and $y_r$, velocity $ v_x $ and $ v_y $, intensity $ \sigma $, and false alarm probability of cluster $ \gamma_{fp} $ are considered to determine $\mathbf{H}^{radar}(x,y)$ in~(\ref{eq-rh-3}). Therefore  we set the channel number of radar heatmaps as six. The mean value is denoted by radar position measurement $p_x$ and $ p_y $ to construct radar heatmaps. The standard deviation $\Sigma$ is a function that combines a fixed kernel size $ \tau$, and the value  of radar measurements as in~(\ref{eq-rh-1}). If the current channel reflects position and velocity, root mean square is considered to determine $ \Sigma $. Otherwise, it is defined by the value itself. Besides, the amplitude value $ M $ of each heatmap is denoted as~(\ref{eq-rh-2}).
\begin{equation}
	\label{eq-rh-3}
	\mathbf{H}^{radar}(x,y,a)= \frac{1}{M(a)}
	\begin{cases}
		e^{- \frac{(x-p_x)^2+(y-p_y)^2)}{2\Sigma(a)}}& \text{ $\left| x-p_x \right| \in 3\Sigma_x$, }\\ 
		&\ \text{$\left| y-p_y \right| \in 3\Sigma_y $ } \\
		0& \text{otherwise}
	\end{cases}
\end{equation}

\begin{equation}
	\label{eq-rh-1}
	\Sigma(a) = 
	\begin{cases}
		diag\left( max\left( a_{rms},\tau\right)\right)  & \text{ $a\in\left\lbrace x_r,y_r,v_x,v_y\right\rbrace$}\\ 
		diag\left( max(a, \tau)\right) &\  \text{$ a\in\left\lbrace \sigma, \gamma_{fp} \right\rbrace $}
	\end{cases}
\end{equation}

\begin{equation}
	\label{eq-rh-2}
	M(a) =(2\pi)\left| \Sigma(a)\right| ^{1/2}
\end{equation}

After getting $ \mathbf{H}^{PF} $ and $  \mathbf{H}^{radar} $, we use multiply operation to fuse the two heatmaps. With $ 1\times1 $ convolution kernel, the $ \mathbf{H}^{RF} $ is obtained which is the same resolution as $ \mathbf{H}^{PF} $. Based on the above operations, ROI-fusion is accomplished.
\subsection{Detection head design}\label{subs-head}
Our detection head is designed to regress the prediction results of the whole fusion model. As we realize an anchor-free detection model, we use a share convolution layer to calculate the regression losses and obtain final outputs. The losses are jointly optimized under BEV.\par
As for the loss design of our fusion model, there are two branches consisting of object classification and location regression, termed $L_{cls}$ and $L_{reg}$ respectively. The total training loss for the whole task is constructed as (\ref{eq-loss}), where $ \beta $ denotes the weight of $  L_{reg}  $.
\begin{equation}
	\label{eq-loss}
	L_{total} = \beta L_{reg} + L_{cls}
\end{equation}
\par
Under BEV, features of foreground objects are relatively sparse compared with background information. Therefore, we utilize focal loss $L_{cls}$ to deal with the unbalanced sample classification. Ground truth information is splatted onto heatmaps  $\mathbf{H}^{GT}\in[0,1]^{\frac{W}{R}\times\frac{H}{R}\times C} $ of $C$ classes in a similar way using a Gaussian kernel
$ \mathbf{H}^{GT}= \exp\left( \left( ( x-\bar{p}_x )^2 +( y-\bar{p}_y )^2\right)  /2\sigma_p^2\right)   $. After the former process, all the features are splatted as feature heatmaps $ \mathbf{H}^{RF}$ under BEV. Then another convolution layer is used to get prediction heatmaps $ \mathbf{H}^{Pred}$ from $ \mathbf{H}^{RF}$ in $C$ channels. In order to do continuous calculations instead of discrete ones, we use another Gaussian function $ \mathcal{G}\left( H^{GT}\right) $ to present the probability of objects. Then, the category loss is calculated as
\par
\begin{equation}
	\begin{aligned}
		L_{cls} = -\frac{1}{N}\sum_{xyc}& \log \left(\mathbf{H}^{Pred}\right) \left( 1-\mathbf{H}^{Pred}\right) ^{\alpha}\mathcal{G}\left( \mathbf{H}^{GT}\right) \\
		&+ \log \left( 1-\mathbf{H}^{Pred}\right) \left( \mathbf{H}^{Pred}\right) ^{\alpha}\left( 1-\mathcal{G}\left( \mathbf{H}^{GT}\right)^{\gamma} \right) 
	\end{aligned}
\end{equation}
where $\alpha$ and $\gamma$ are the constants in focal loss design. Here we set $\alpha=2$ and $\gamma=4$. $ C $ is the number of categories.\par 
As for calculating the loss $L_{reg}$  of 3D information, including 3D location towards $ x$, $ y$, $ z$ position, width $ w $, length $ l $ and height $ h $ of object bounding boxes, velocitiy $ v_x $, $ v_y$, and heading angle $ \theta $ of objects, we follow the L1 loss to regress position offset $ \hat{O}_p $, dimensions, velocities as $ L_{off} $, $ L_{dim} $, and $ L_{vel} $. We use the softmax function to calculate orientation loss $ L_{rot}$, which is encoded with two bins $b$ and eight scalars. Then the Binary Cross Entropy~(BCE) loss is used to calculate attribute loss $ L_{att}  $ of attributes $a$.
\begin{equation}
	\begin{aligned}
		L_{reg} = &\lambda_{off}L_{off}+ \lambda_{dim}L_{dim} + \lambda_{vel}L_{vel} +\lambda_{rot}L_{rot} +\lambda_{att}L_{att}\\
		L_{off} = &\frac{1}{N_T}\sum_{k=1}^{N_T}\sum_{p}\left|\hat{O}_{pk}-(\frac{p_k}{R}-\tilde{p_k})\right|, p = p_x,p_y,p_z \\
		L_{dim} = &\frac{1}{N_T}\sum_{k=1}^{N_T}\left|s_k-\hat{s}_k\right|, s = w,l,h\\
		L_{vel} = &\frac{1}{N_T}\sum_{k=1}^{N_T}\left|v_k-\hat{v}_k\right|, v = v_x,v_y\\
		L_{rot} = &\frac{1}{N_T}\sum_{k=1}^{N_T}\left[ softmax(\hat{b}_k) + \frac{-1}{N_\theta}\sum_{i=1}^{N_\theta} \cos\left(\theta-b_i-\Delta\theta_i \right)\right]  \\
		L_{att} = &\frac{1}{N_T}\sum_{k=1}^{N_T}\frac{1}{N_A}\sum_{i=1}^{N_A}-w_i\left|a_k^i\log\hat{a}_k^i+(1-a_k^i)\log(1-\hat{a}_k^i)\right|\\
	\end{aligned}
\end{equation}
where $ N_T $ is the number of training samples,   $ N_{\theta} $ is the number of bins that cover the heading angle $\theta$, $ b_i $ is the center angle of the bin $i$, and $\Delta\theta$ is the angular offset that needs to be applied to the center of bin $i$. $ N_A $ is the number of attributes in driving scenes. \par
\section{Experiment}\label{sec-experiment}
\subsection{Dataset and Evaluation metrics}
Experiments are implemented on the large-scale public dataset nuScenes~\cite{caesar2020nuscenes} for research on autonomous driving. Data from 1000 driving scenes are collected in Boston and Singapore. Approximately 1.4M camera images, 390k LIDAR sweeps, 1.4M RADAR sweeps, and 1.4M object bounding boxes in 40k keyframes are recorded. Except for typical scenes, special scenarios such as rainy days and nights are also included in this dataset.\par
We use the metrics which are designed for the nuScenes data. There are three kinds of official metrics to evaluate detection performance. First, the Average Precision metric is assessed by mean Average Precision~(mAP), which is calculated for measuring the precision and recall of detection methods. However, in nuScenes metrics, it is not defined based on the Intersection over Union~(IOU), but the match by 2D center distance on the ground plane. Second, True Positive~(TP) metrics are used to evaluate the multi-aspect precision, including Average Translation Error (ATE), Average Scale Error (ASE), Average Orientation Error (AOE), Average
Velocity Error (AVE), and Average Attribute Error (AAE) of detection results. Third, the nuScenes detection score~(NDS) is designed to indicate  the all-sided detection performance, which considers mAP and the regression quality in terms of box location, size, orientation, attributes, and velocity. As in~(\ref{eq-metric}), the above metrics are calculated over the distance matching threshold of $ \mathbb{D}=\left\lbrace 0.5,1,2,4\right\rbrace $ and the set of ten classes $ \mathbb{C} $.\par
\begin{equation}
	\begin{aligned}
		mAP &=\frac{1}{\left|\mathbb{C}\right| \left|\mathbb{D}\right|} \sum_{c\in \mathbb{C}}\sum_{d\in \mathbb{D}}AP_{c,d}\\
		mTP &= \frac{1}{\left|\mathbb{C}\right|  }\sum_{c\in \mathbb{C}} TP_c \\
		NDS &= \frac{1}{10}\left[ 5mAP + \sum_{mTP\in TP}\left( 1-\min \left( 1, mTP\right) \right) \right] 
	\end{aligned}
	\label{eq-metric}
\end{equation}

\begin{figure*}[!htb]
	\centering
	\includegraphics[scale=0.6]{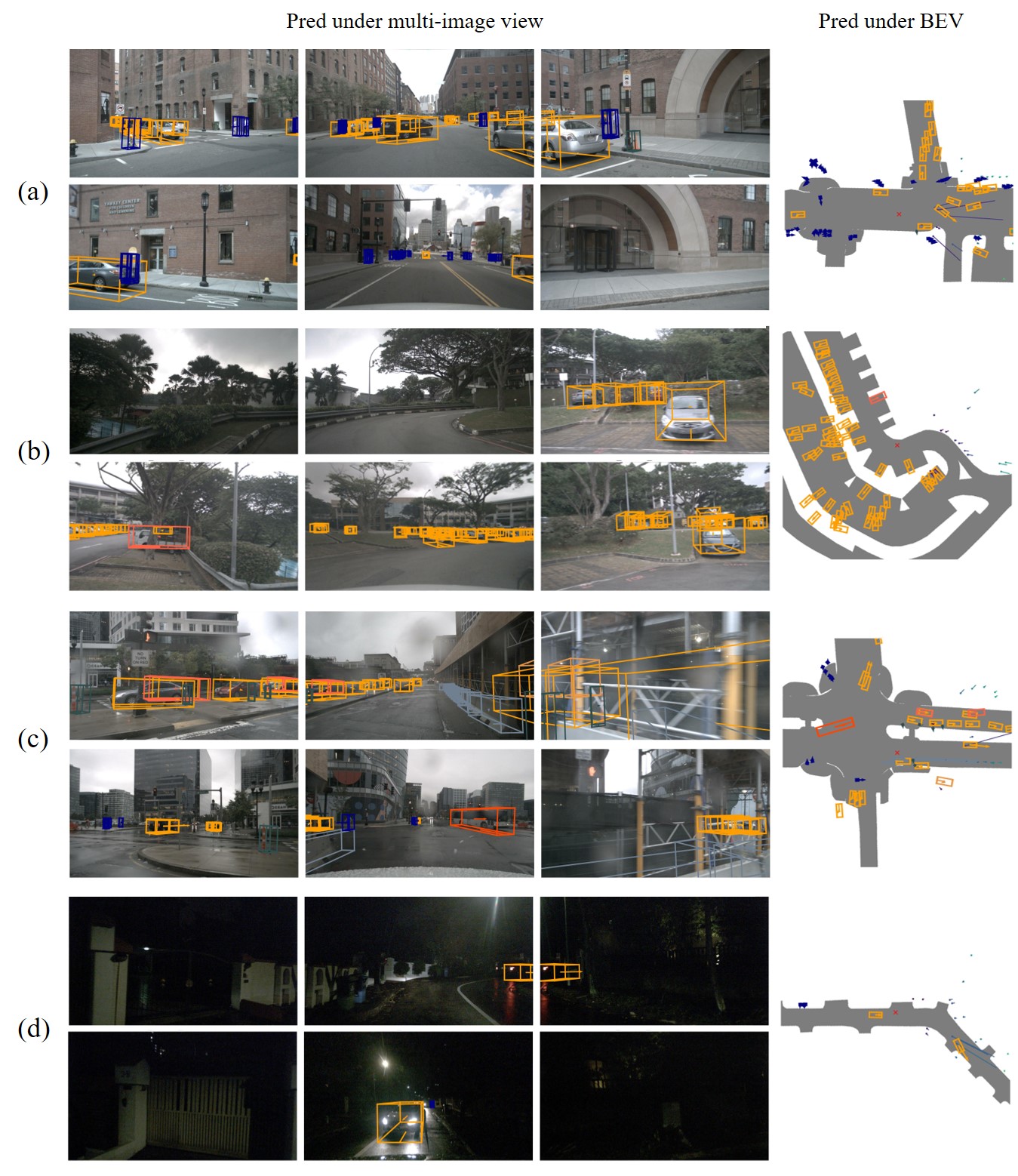}
	\caption{3D detection results on nuScenes dataset of RCBEV with R-C fusion. The pictures from left to right are detection results under multi-image view and BEV. The scenes are on city roads, parking lots, rainy days, and nights.}
	\label{fig-demo}
\end{figure*}
\begin{table*}[!h]
	\centering
	\caption{Performance comparisons for 3D object detection on nuScenes dataset.}
	\begin{threeparttable}
		\vspace{0.2mm}
		\scalebox{1.0}{
			\begin{tabular}{cccccccccc}
				\toprule
				\toprule
				\multicolumn{1}{c}{ \textbf{Method}} & \multicolumn{1}{c}{\textbf{Modality}} &\multicolumn{1}{c}{\textbf{data}} & \multicolumn{7}{c}{\textbf{Metric}}  \\
				\cmidrule(lr){4-10}\\
				& & & NDS $\uparrow$ & mAP$\uparrow$ & mATE(m) $\downarrow$ & mASE(1-IoU) $\downarrow$ &mAOE(rad) $\downarrow$ &mAVE(m/s) $\downarrow$  & mAAE(1-acc) $\downarrow$\\ 
				\midrule
				BEVDet     &Camera    &val      & 0.403    & 0.308    &  0.665   &  0.273   & 0.533 & 0.829 & 0.205 \\
				BEVDet4d   &Camera    &val      & 0.476    & 0.338    &  0.671   &  0.274   & 0.469 & \textbf{0.337} & 0.185 \\
				CenterFusion  &Camera+Radar  &val  &  0.453   &  0.332   &  0.649   &  \textbf{0.263}   & 0.535  & 0.540 &\textbf{0.142}  \\
				RCBEV(Ours) &Camera+Radar  &val   &0.482 &0.377 &0.534 &0.271 &0.558 & 0.493 & 0.209   \\
				RCBEV4d(Ours) &Camera+Radar  &val   &\textbf{0.497}  &\textbf{0.381} &\textbf{0.526 }&0.272 &\textbf{0.445} &0.465 &0.185\\
				\midrule
				mm-fusion &Camera+Radar &test &0.483 &0.371 &0.628 &\textbf{0.250} &\textbf{0.513}&\textbf{0.536}&\textbf{0.095}\\
				CenterFusion &Camera+Radar &test &0.449 &0.326 &0.631 &0.261&0.516 &0.614&0.115\\
				RCBEV(Ours)  &Camera+Radar &test &\textbf{0.486} &\textbf{0.406} &\textbf{0.484} &0.257 &0.587&0.702&0.140\\
				\bottomrule
				\bottomrule
			\end{tabular}
		}
		\begin{tablenotes} 
			\footnotesize 
			\item $\uparrow$ denotes the higher the value is, the better the performance is.$\downarrow$ denotes the lower the value is, the better the performance is.
			\item Numbers in bold represent the best performance under this criteria.      
		\end{tablenotes}  
	\end{threeparttable}
	\label{tab-overall}
\end{table*}

\begin{table*}[h]
	\centering
	\caption{Per-class performance comparisons for 3D object detection on nuScenes dataset.}
	\begin{threeparttable}
		\vspace{0.2mm}
		\scalebox{1.0}{
			\begin{tabular}{ccccccccccccc}
				\toprule
				\toprule
				\multicolumn{1}{c}{ \textbf{Method}} & \multicolumn{1}{c}{\textbf{Modality}} &\multicolumn{1}{c}{\textbf{data}} & \multicolumn{10}{c}{\textbf{mAP$ \uparrow $}}  \\
				\cmidrule(lr){4-13}\\
				& & & Car & Truck & Bus & Trailer & Const. & Pedest. & Motor. &Bicycle &Traff. &Barrier\\ 
				\midrule
				BEVDet     &Camera    &val      & 0.508    & 0.222    &  0.311   &  0.150   & 0.073  & 0.336  & 0.262 & 0.213  & 0.506  & 0.502  \\
				BEVDet4d     &Camera    &val      & 0.536    & 0.247    &  0.335   &  \textbf{0.160}    & 0.079  & 0.384  & 0.292 & \textbf{0.278}  & 0.546  & 0.522  \\
				CenterFusion    &Camera+Radar    & val  & 0.524 &  0.265   &  0.362   &  0.154   & 0.055  & 0.389 & 0.305 & 0.229  & \textbf{0.563}  & 0.470  \\
				RCBEV(Ours)&Camera+Radar  &val   &0.675 &\textbf{0.333} &\textbf{0.410} & 0.152 & \textbf{0.112}  &0.419   & 0.325 & 0.248 & 0.547 & 0.545  \\
				RCBEV4d(Ours) &Camera+Radar  &val   &\textbf{0.683} &0.323 &0.369 &0.148 &0.108 &\textbf{0.443} &\textbf{0.357 }&0.270 &0.552 &\textbf{0.557} \\
				\midrule
				mm-fusion &Camera+Radar &test &0.603 &0.301 &0.260 &0.225 &0.112 &\textbf{0.436} &0.399 &\textbf{0.298}&0.556 &0.518\\
				CenterFusion  &Camera+Radar &test &0.509  &0.258  &0.234  &0.235 & 0.077&0.370 &0.314& 0.201&0.575 &0.484\\
				RCBEV(Ours)  &Camera+Radar &test &\textbf{0.663}  &\textbf{0.332}  &\textbf{0.316}  &\textbf{0.332} &\textbf{ 0.166} & 0.412&\textbf{0.416} & 0.265&\textbf{0.611} &\textbf{0.551}\\
				\bottomrule
				\bottomrule
			\end{tabular}
		}
		\begin{tablenotes} 
			\footnotesize 
			\item Ten categories include 'Car', 'Truck',  'Bus', 'Trailer',  'Construction vehicle', 'Pedestrian','Motorcycle', 'Bicycle', 'Traffic cone', 'Barrier'.
		\end{tablenotes}  
	\end{threeparttable}
	\label{tab-perclass}
\end{table*}

\subsection{Implementation details}
This paper implemented RCBEV on MMDetection3D~\cite{contributors2020mmdetection3d}, a well-conducted platform for 3D perception tasks.  The model is trained on 4 NVIDIA GeForce GTX 3090 GPUs with 24GB memory.  The AdamW optimizer~\cite{loshchilov2017decoupled} is adopted during the training process to obtain optimized parameters. The learning rate is initialized to $2e-4$. The training epoch is set as 20. And the total batch size is 32. Besides, we exploit CBGS~\cite{zhu2019class}, a class-balanced data grouping and sampling strategy during training, for an ideal result. The vision-branch backbone network of RCBEV is first pre-trained on the nuScenes dataset. \par
During training and testing, six cameras and five radars mounted on the vehicles are utilized for a more broad field of view.  The image resolution is reduced from the original $1600\times900$ pixels to $704\times256$ pixels for a balance between precision and computational efficiency. We accumulate ten sweeps, which is a commonly used time window length to increase radar point cloud density. To further realize the data augmentation, random right-left or forward-backward flipping (with a probability of 0.5), random rotation (from -0.3925rad to 0.3925rad under LiDAR coordinate system), and random scaling (from 0.95 to 1.05 times of point range) are implemented on radar point clouds. Voxel size is set to [0.1, 0.1, 0.2] meters towards the [x, y, z] directions.\par
\subsection{Overall result analysis}
The visible results of our method are shown in Fig.~\ref{fig-demo}. Four representative scenarios are selected from the nuScenes dataset.  At the left part of the sub-figures, detection results are marked on the multiple image views. On the right parts, our detection results are shown under BEV. The different colors of bounding boxes denote different  object categories.\par
As for the quantitative analysis, we compare our radar and camera fusion 3D detection work RCBEV with the baseline  camera-based method BEVDet~\cite{huang2021bevdet}, its improved version BEVDet4d~\cite{huang2022bevdet4d}, and the SOTA camera-radar fusion models like CenterFusion~\cite{nabati2021centerfusion} on the nuScenes benchmark. As nuScenes is split as training, validation, and testing dataset, the training part is used for model learning, and the validation and testing parts are used for model evaluation. The sub-datasets do not coincide with each other.\par   
Out of fairness and rigor, when evaluating our model on the validation set, the relevant factors which influence the performance are kept as constant as possible. Consequently, the image size, and the backbone of the image branch are kept the same among BEVDet, BEVDet4d, RCBEV, and RCBEV4d. Here BEVDet and RCBEV only use a single image frame, while BEVDet4d and RCBEV4d exploit temporal cues from images as another data source. The image backbone and image neck are adopted as Swin-Transformer and FPN. When evaluating our model on the testing set, we upload our detection results and obtain the evaluation results from the nuScenes evaluation server. The highest results on the leaderboard under the camera-radar modality, including the results of mm-fusion, CenterFusion, and our work, are provided in Table~\ref{tab-overall}. \par
From Table~\ref{tab-overall}, we can see direct comparisons of different algorithms.  Compared with the other methods, our method attains better performance in nearly all criteria. RCBEV outperforms BEVDet with 6.9\% mAP and 7.9\% NDS. Even when comparing with BEVDet4d, which imports temporal features of multiple images, RCBEV still overcomes with 3.9\% mAP and 0.6\% NDS.  After exploring sequential information, RCBEV4d overcomes BEVDet4d with 4.3\% mAP and 2.1\% NDS. Then when we focus on the regression precision of location, rotation, and dimensions of objects, we find after utilizing radar information, the location error, dimension error, and velocity error of detection results, which are denoted by mATE, mASE, and mAVE, achieve a significant reduction compared with BEVDet. \par 
Besides, we receive the best performance on 3D detection task of camera-radar modality. On the validation set, RCBEV  exceeds Centerfusion, the most representative public work on camera-radar fusion 3D detection  with 4.5\% mAP and 2.9\% NDS.  RCBEV4d exceeds Centerfusion with 4.9\% mAP and 4.4\% NDS.  On the testing set, RCBEV  receives the highest mAP and NDS scores, with 40.6\%  and 48.6\%. The performance comparison among all the camera-radar fusion works demonstrates the superior performance of the RCBEV.\par
\begin{table*}[!h]
	\centering
	\caption{The robust validation of RCBEV in different lighting and weather conditions.}
	\begin{threeparttable}
		\vspace{0.2mm}
		\scalebox{1.0}{
			\begin{tabular}{cccccccccc}
				\toprule
				\toprule
				& & \multicolumn{2}{c}{Sunny} &\multicolumn{2}{c}{Rainy} & \multicolumn{2}{c}{Day} & \multicolumn{2}{c}{Night} \\
				\cmidrule(lr){3-4}	\cmidrule(lr){5-6}	\cmidrule(lr){7-8}	\cmidrule(lr){9-10}\\
				\textbf{Method}  & \textbf{Modality} & mAP &NDS & mAP & NDS &mAP & NDS &mAP &NDS\\
				\midrule
				BEVDet            &C         &0.304   &0.391     &0.318   &0.462 &0.312 &0.407 &0.134 &0.231 \\
				
				RCBEV(Ours) &C+R    &0.361    &0.466    &0.385   &0.500  &0.371 &0.479 &0.155 &0.237\\
				\midrule
				\multicolumn{2}{c}{Difference Value} &+5.7\%    &+7.5\%  &+6.7\%   &+3.8\%  &+5.9\% &+7.2\% &+2.1\% &+0.6\%\\ 
				\bottomrule
				\bottomrule
			\end{tabular}
		}
		\begin{tablenotes} 
			\footnotesize 
			\item { \qquad \qquad \qquad \qquad Notice that "C" and "R" specify camera and radar modalities, respectively.}
		\end{tablenotes}  
	\end{threeparttable}
	\label{tab-robust}
\end{table*}

\subsection{Evaluation of special cases}
Table~\ref{tab-perclass} displays the per-class performance of several methods mentioned above. As shown, compared with the camera-only method, the accuracy of RCBEV has made apparent improvements in the following categories: car, truck, bus, construction vehicle, pedestrian, motorcycle, traffic cone, and barrier. It is analyzed that radar data gives more objects' geometry constraints, which are helpful to the detection task.\par
Table~\ref{tab-robust} presents the performance of our fusion model and baseline model under different weather and lighting conditions. The metrics mAP and NDS get improved separately under sunny weather, rainy weather, daytime, and night weather. Detection in night and rainy weather is challenging for the camera-only models due to the properties of camera lens. The stable working performance of radar in adverse weather conditions improves the robustness of our fusion model prominently.  \par

\subsection{Ablation study}
In this section, ablation experiments are performed in the validation dataset to further prove the effectiveness of our proposed RCBEV.
\begin{figure*}[!htb]
	\centering
	\includegraphics[width=\linewidth]{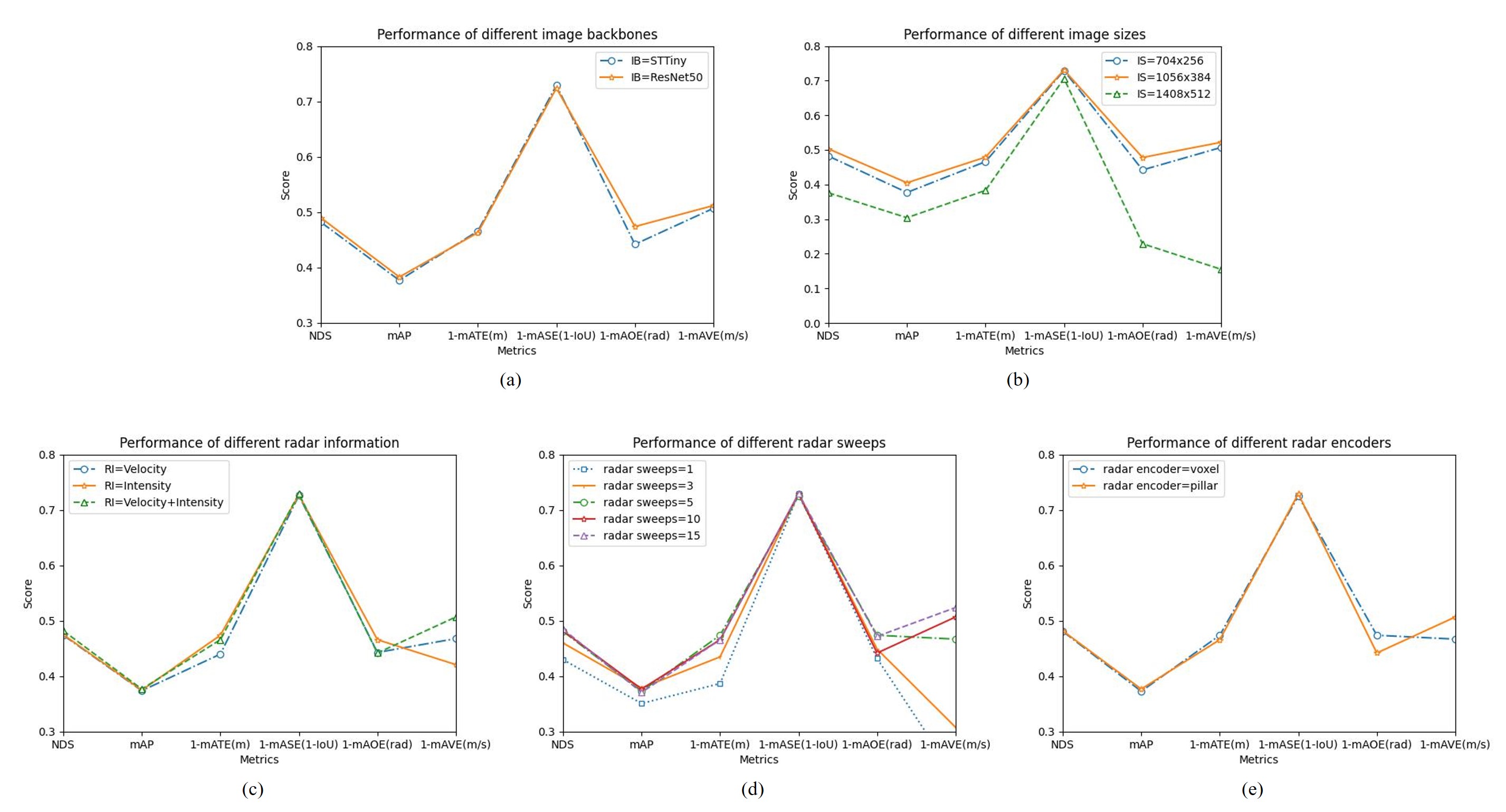}
	\caption{The curves of model performance in the ablation study}
	\label{fig-ablation}
\end{figure*}
We control the variants of RCBEV design as shown in Table~\ref{tab-ablation-ib}-Table~\ref{tab-ablation-re}. The performance curve is visualized in Fig.~\ref{fig-ablation}. \par

By comparing different image backbone networks in Table~\ref{tab-ablation-ib}, the performance of Swin-Transformer and ResNet both get apparent improvements after fusing radar data. The NDS and mAP correspondingly increase (7.9\%, 6.9\%) and (12.3\%, 8.4\%). Hence, it is confirmed that the fusion result can be better than the camera-only result with the same image backbone. There exists a small distinction between the performances of different backbone networks, which may be relevant to the different learning policies adopted and different convergences of models.\par

By comparing different image resolutions in Table~\ref{tab-ablation-is}, we can see that the best performance is carried out with a 1056$ \times $384 image resolution. The highest performance is 50.3\% NDS and 40.5\% mAP.  It is inferred that increasing image size suitably can increase the final performance. However, when exceeding the ideal range, the increase in image size brings difficulty to the training process and harms the detection precision instead.\par

Not only camera-related components but also radar-related components are discussed here. We change the properties which are used with position information to extract features in the radar-branch spatial encoder. The corresponding performance is shown in Table~\ref{tab-ablation-ri}. While both velocity and intensity information is effective, the best performance is obtained by integrating them. Besides, it is indicated that using velocity information can reduce the error of velocity estimation in contrast to not using it. \par

Moreover, we change the time window length when extracting features in the radar-branch temporal encoder, and the result is shown in Table~\ref{tab-ablation-rs}. According to the sampling frequency, 3,5,10,15 sweeps correspond to approximately 0.5s, 1s, 1.5s, and 2.5s. We also experiment with single radar input. It is figured that  more sweeps are good for the overall performance, especially velocity estimation. As for the mAP metric, it is indicated that when the time window size is 3, mAP gets the highest score. Comparing with CenterFusion, which also uses multi-sweep accumulation, our RCBEV overcomes  CenterFusion with  4.7\% mAP and 0.7\% NDS when time window sizes equal 3. \par
 
The performances of different radar point encoder forms are shown in Table~\ref{tab-ablation-re}. Both models prevail over the camera-only methods. Furthermore, the performance of the voxel encoder is better than the pillar encoder. \par

From the comparisons of the above components, we can see that both radar-related and camera-related components have important influences on the final fusion performance. When choosing an appropriate feature extractor with suitable parameters, the fusion result surpass the single-sensor model for certain.\par


\begin{table}[!htbp]
	\centering
	\caption{ablation experiments of different image backbones of RCBEV on nuScenes}
	\begin{threeparttable}
		\vspace{0.2mm}
		\scalebox{0.8}{
			\begin{tabular}{cccccccc}
				\toprule
				\toprule
				Backbone &Modality & NDS $\uparrow$ & mAP $\uparrow$ & mATE $\downarrow$ & mASE $\downarrow$ &mAOE $\downarrow$ &mAVE $\downarrow$ \\ 
				\midrule
				STTiny &C &0.403    & 0.308  &  0.665   &  0.273 & 0.533 & 0.829   \\
				Resnet50  &C &0.377  &0.299 &0.734  &0.273 &0.573 &0.907  \\
				\midrule
				STTiny &R+C &0.482  &0.377  &0.534 &0.271  &0.558 &0.493   \\
				Resnet50  &R+C &0.490  &0.383  &0.537  &0.276 &0.526 &0.488  \\
				\bottomrule
				\bottomrule
			\end{tabular}
		}
	\end{threeparttable}
	\label{tab-ablation-ib}
\end{table}

\begin{table}[!htbp]
	\centering
	\caption{ablation experiments of different image sizes of RCBEV on nuScenes}
	\begin{threeparttable}
		\vspace{0.2mm}
		\scalebox{0.8}{
			\begin{tabular}{ccccccc}
				\toprule
				\toprule
				Image size & NDS $\uparrow$ & mAP $\uparrow$ & mATE $\downarrow$ & mASE $\downarrow$ &mAOE $\downarrow$ &mAVE $\downarrow$ \\ 
				\midrule
				704$\times$256  &0.482  &0.377  &0.534 &0.271  &0.558 &0.493   \\
				1056$\times$384  &0.503  &0.405  &0.521  &0.269 &0.522 &0.478  \\
				1408$\times$512  &0.376  &0.304  &0.617  &0.294 &0.771 &0.845  \\
				\bottomrule
				\bottomrule
			\end{tabular}
		}
	\end{threeparttable}
	\label{tab-ablation-is}
\end{table}

\begin{table}[!htbp]
	\centering
	\caption{ablation experiments of radar information of RCBEV on nuScenes}
	\begin{threeparttable}
		\vspace{0.2mm}
		\scalebox{0.8}{
			\begin{tabular}{ccccccc}
				\toprule
				\toprule
				radar data & NDS $\uparrow$ & mAP $\uparrow$ & mATE $\downarrow$ & mASE $\downarrow$ &mAOE $\downarrow$ &mAVE $\downarrow$ \\ 
				\midrule
				V &0.474  & 0.374 &0.560 &0.274  &0.557 & 0.532  \\
				I &0.475  &0.375  &0.526  &0.273 &0.534 &0.579  \\
				V+I &0.482  &0.377  &0.534 &0.271  &0.558 &0.493   \\
				\bottomrule
				\bottomrule
			\end{tabular}
		}
	\end{threeparttable}
	\label{tab-ablation-ri}
\end{table}

\begin{table}[!htbp]
	\centering
	\caption{ablation experiments of radar sweeps of RCBEV on nuScenes}
	\begin{threeparttable}
		\vspace{0.2mm}
		\scalebox{0.8}{
			\begin{tabular}{ccccccc}
				\toprule
				\toprule
				radar sweeps & NDS $\uparrow$ & mAP $\uparrow$ & mATE $\downarrow$ & mASE $\downarrow$ &mAOE $\downarrow$ &mAVE $\downarrow$ \\ 
				\midrule
				1 &0.430  &0.351  &0.613 &0.271  &0.568 &0.782   \\
				3 &0.460  &0.379  &0.565 &0.270  &0.551 &0.692    \\
				5 &0.481  &0.373  &0.526 &0.275  &0.526 &0.533   \\
				10 &0.482  &0.377  &0.534 &0.271  &0.558 &0.493    \\
				15 &0.485  &0.371  &0.534  &0.271 &0.528& 0.476  \\
				\bottomrule
				\bottomrule
			\end{tabular}
		}
	\end{threeparttable}
	\label{tab-ablation-rs}
\end{table}

\begin{table}[!htbp]
	\centering
	\caption{ablation experiments of different radar encoder of RCBEV on nuScenes}
	\begin{threeparttable}
		\vspace{0.2mm}
		\scalebox{0.8}{
			\begin{tabular}{ccccccc}
				\toprule
				\toprule
				radar encoder  & NDS $\uparrow$ & mAP $\uparrow$ & mATE $\downarrow$ & mASE $\downarrow$ &mAOE $\downarrow$ &mAVE $\downarrow$ \\ 
				\midrule
				voxel  &0.482  &0.377  &0.534 &0.271  &0.558 &0.493   \\
				pillar  &0.477  &0.376  & 0.535 &0.273 & 0.577&0.515  \\
				\bottomrule
				\bottomrule
			\end{tabular}
		}
	\end{threeparttable}
	\label{tab-ablation-re}
\end{table}

\section{Conclusion}\label{sec-conclusion}
This paper proposes a novel radar-camera feature fusion 3D object detection method called RCBEV. In contrast to the other fusion detection method with radar and camera, RCBEV bridges the view disparity between radar and camera features and realizes an efficient feature fusion system under the top-down view instead of the front view. Besides, aiming at the tough nut from the sparsity and clutter of radar data, a radar temporal-spatial encoder  is designed to maximize the efficient feature extraction from radar points. Then our two-stage fusion method gives us a more sufficient information interaction during the fusion course to enhance the regression result. The model performance is validated on the large-scale benchmark nuScenes. Extensive experiments demonstrate that our method can successfully optimize the detection result compared with the baseline of camera-only methods. What's more, the best detection performance has been achieved by the RCBEV model among all the camera-radar fusion methods up to now. \par 
RCBEV verifies that by covering the view disparity between image and radar data under the top-down view, obvious improvements in accuracy and robustness are achieved. However, the current work still stays on the dynamic object perception task. Static elements in the driving environment, such as lane lines, and lamp poles, which can be regarded as landmarks and contribute to  location and mapping in autonomous driving, have not been considered yet.  As for future work, we will study how to utilize our fusion model to realize a unified perception of dynamic and static elements in driving scenes to reconstruct an  integrated 3D environment.\par
\section*{Acknowledgments}
This work was supported by the National Natural Science Foundation of China (U1864203, 52102396) and partly funded by the China Postdoctoral Science Foundation (2021M701897).

\bibliographystyle{IEEEtran}
\bibliography{thbibfile}
%
%

\section*{Biography Section}
%
%
\begin{IEEEbiography}[{\includegraphics[width=1in,height=1.25in,clip,keepaspectratio]{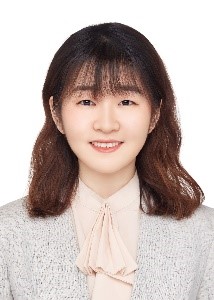}}]{Taohua Zhou} received the B.S. degree from Department of Automation, Tsinghua University, Beijing, China in 2018. She is currently working toward the Doctor degree at the School of Vehicle and Mobility, Tsinghua University, Beijing, China. Her research interests include object detection and tracking, information fusion, environmental perception of autonomous driving and vehicle-infrastructure cooperative perception.
\end{IEEEbiography}

\begin{IEEEbiography}[{\includegraphics[width=1in,height=1.25in,clip,keepaspectratio]{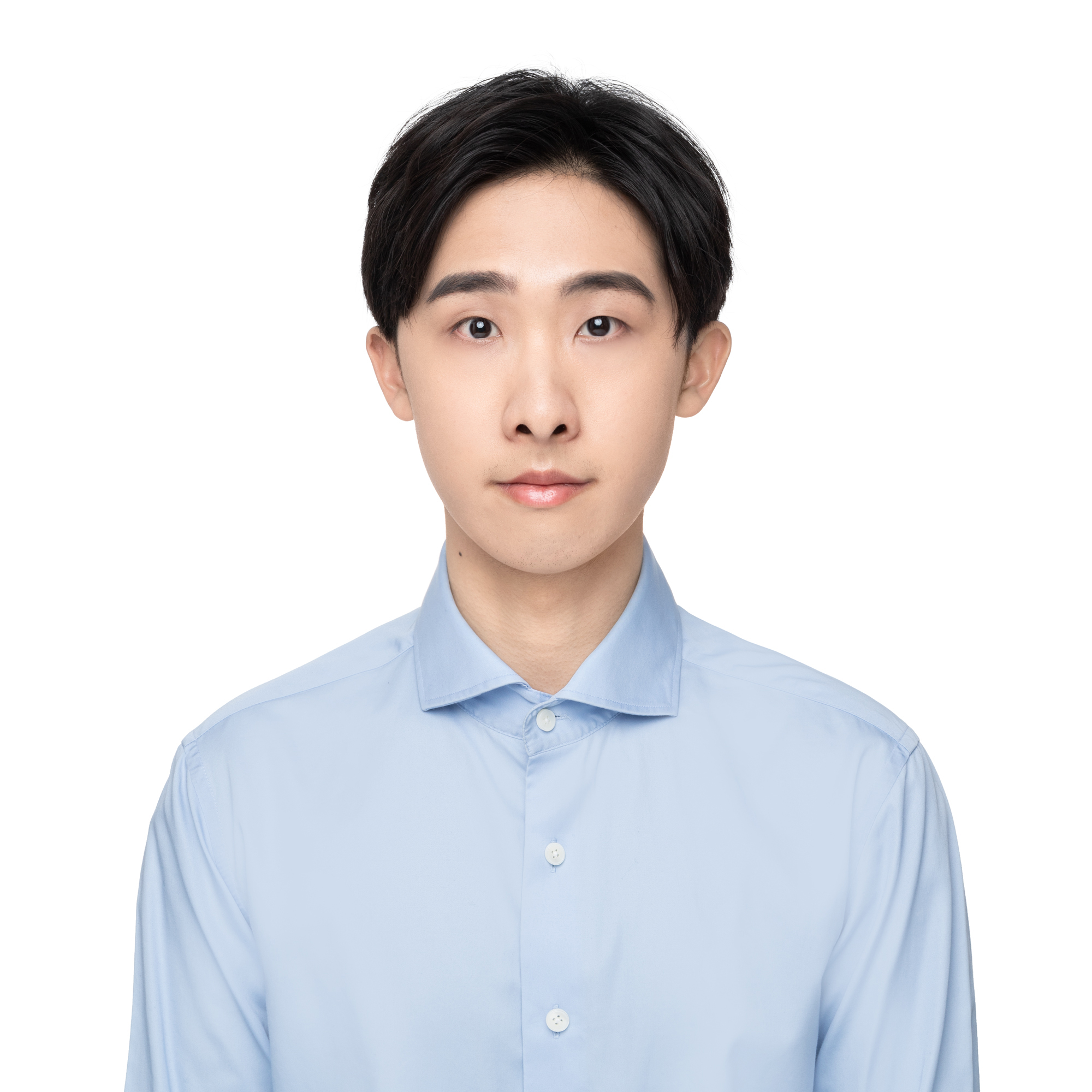}}]{Yining Shi} received the B.S. degree in School of Automotive Engineering from Tsinghua University, Beijing, China in 2021. He is currently working toward the Ph.D. degree at the School of Vehicle and Mobility in Tsinghua University, Beijing, China. His research interests include multi-modality 3D object detection and tracking, as well as grid-centric segmentation and motion prediction for autonomous driving.
\end{IEEEbiography}

\begin{IEEEbiography}[{\includegraphics[width=1in,height=1.25in,clip,keepaspectratio]{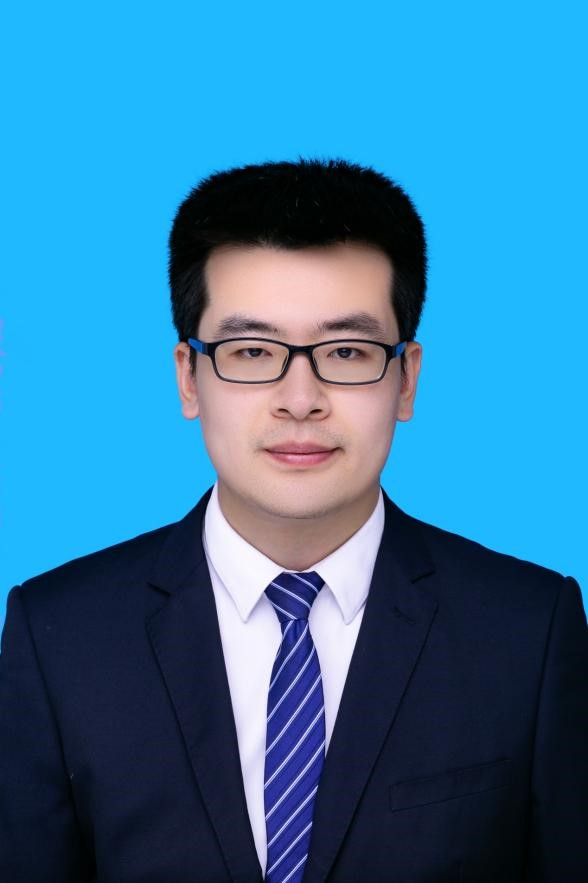}}]{Junjie Chen } received the Ph.D. degree in traffic information engineering and control from the Beijing Jiaotong University in 2020. He was a Research Assistant at Carnegie Mellon University (CMU), Pittsburgh, PA, USA, from 2018 to 2020. He currently holds a post-doctoral position at the School of Vehicle and Mobility of Tsinghua University, Beijing, China. His research interests include nonparametric Bayesian learning, platoon operation control, and recognition and application of human driving characteristics.
\end{IEEEbiography}

\begin{IEEEbiography}
	[{\includegraphics[width=1in,height=1.25in,clip,keepaspectratio]{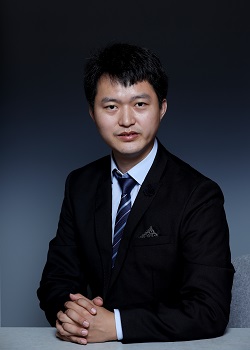}}]{Kun Jiang}
	received the B.S. degree in mechanical and automation engineering from Shanghai Jiao Tong University, Shanghai, China in 2011. Then he received the Master degree in mechatronics system and the Ph.D. degree in information and systems technologies from University of Technology of Compi\`egne (UTC), Compi\`egne, France, in 2013 and 2016, respectively. He is currently an assistant research professor at Tsinghua University, Beijing, China. His research interests include autonomous vehicles, high precision digital map, and sensor fusion.
\end{IEEEbiography}

\begin{IEEEbiography}
[{\includegraphics[width=1in,height=1.25in,clip,keepaspectratio]{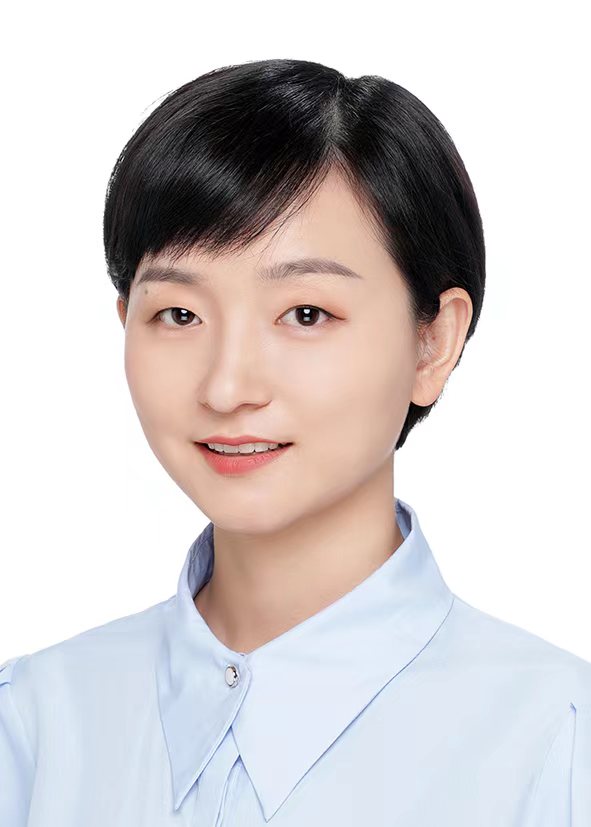}}]{Mengmeng Yang} received the Ph.D. degree in Photogrammetry and Remote Sensing from Wuhan University, Wuhan, China in 2018. She is currently an assistant research professor at Tsinghua University, Beijing, China. Her research interests include autonomous vehicles, high precision digital map, and sensor fusion. 
\end{IEEEbiography}

\begin{IEEEbiography}[{\includegraphics[width=1in,height=1.25in,clip,keepaspectratio]{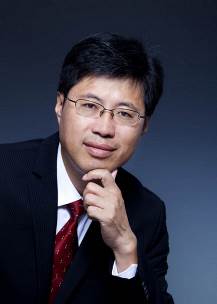}}]{Diange Yang} received his Ph.D. in Automotive Engineering from Tsinghua University in 2001. He is now a Professor at the school of vehicle and mobility at Tsinghua University. He currently serves as the director of the Tsinghua University Development \& Planning Division. His research interests include autonomous driving, environment perception, and HD map. He has more than 180 technical publications and 100 patents. He received the National Technology Invention Award and Science and Technology Progress Award in 2010, 2013, 2018, and the Special Prize for Progress in Science and Technology of China Automobile Industry in 2019. 
\end{IEEEbiography}


\vfill

\end{document}